\documentclass{edm_article}

\usepackage{amsthm}
\usepackage{booktabs}
\usepackage{xspace}

\usepackage{xcolor}
\usepackage{ifthen}
\newboolean{include-notes}
\setboolean{include-notes}{true}
\newcommand{\nb}[3]{\ifthenelse{\boolean{include-notes}}{{\colorbox{#2}{\bfseries\sffamily\scriptsize\textcolor{white}{#1}}}{\ \textcolor{#2}{\sf\small\textit{#3}}}}{}}

\newtheorem{definition}{Definition}
\newtheorem{observation}{Observation}

\renewcommand{\c}{\ensuremath{\mathbf{c}}}
\newcommand{\x}{\ensuremath{\mathbf{x}}}

\newcommand{\calD}{\ensuremath{\mathcal{D}}}

\newcommand{\calH}{\ensuremath{\mathcal{H}}}
\newcommand{\calI}{\ensuremath{\mathcal{I}}}

\newcommand{\calS}{\ensuremath{\mathcal{S}}}

\newcommand{\calZ}{\ensuremath{\mathcal{Z}}}

\newcommand{\history}{\ensuremath{\calH}}
\newcommand{\items}{\ensuremath{\calI}}

\newcommand{\terminate}{\ensuremath{\top}}
\newcommand{\threshold}{\ensuremath{\tau}}

\newcommand{\nskills}{\ensuremath{K}}
\newcommand{\policies}{\ensuremath{\Pi}}
\newcommand{\skills}{\ensuremath{\calZ}}
\newcommand{\students}{\ensuremath{\calS}}

\newcommand{\BBKT}{B\textsuperscript{2}KT\xspace}

\newcommand{\expectimaxPolicy}[1]{\textsc{Expectimax($#1$)}}

\newcommand{\thresholdPolicy}[1]{\textsc{Threshold($#1$)}}

\begin{document}

\title{Equity and Fairness of Bayesian Knowledge Tracing}
%\subtitle{[Extended Abstract]
%\titlenote{A full version of this paper is available as
%\textit{Author's Guide to Preparing ACM SIG Proceedings Using
%\LaTeX$2_\epsilon$\ and BibTeX} at
% Submissions for EDM are double-blind: please do not include any
% author names or affiliations in the submission. 
% Anonymous authors:
\numberofauthors{3}
\author{
% % You can go ahead and credit any number of authors here,
% % e.g. one 'row of three' or two rows (consisting of one row of three
% % and a second row of one, two or three).
% %
% % The command \alignauthor (no curly braces needed) should
% % precede each author name, affiliation/snail-mail address and
% % e-mail address. Additionally, tag each line of
% % affiliation/address with \affaddr, and tag the
% % e-mail address with \email.
% %
% % 1st. author
\alignauthor
 Sebastian Tschiatschek\\
        \affaddr{University of Vienna}\\
        \affaddr{Faculty of Computer Science}\\
        \affaddr{{Research Network Data Science}}\\
        \affaddr{Vienna, Austria}\\
        %\email{sebastian.tschiatschek@univie.ac.at}
%       
% 2nd. author
\alignauthor
 Maria Knobelsdorf\\
        \affaddr{University of Vienna}\\
        \affaddr{Faculty of Computer Science}\\
        \affaddr{Computer Science Education}\\
        \affaddr{Vienna, Austria}\\
        %\email{maria.knobelsdorf@univie.ac.at}
% 3rd. author
\alignauthor 
 Adish Singla\\
       \affaddr{Max-Planck Institute for Software Systems}\\
       \affaddr{Saarbrücken, Germany}\\
       %\email{adishs@mpi-sws.org}
% \and  % use '\and' if you need 'another row' of author names
% % 4th. author
% \alignauthor Lawrence P. Leipuner\\
%        \affaddr{Brookhaven Laboratories}\\
%        \affaddr{Brookhaven National Lab}\\
%        \affaddr{P.O. Box 5000}\\
%        \email{lleipuner@researchlabs.org}
% % 5th. author
% \alignauthor Sean Fogarty\\
%        \affaddr{NASA Ames Research Center}\\
%        \affaddr{Moffett Field}\\
%        \affaddr{California 94035}\\
%        \email{fogartys@amesres.org}
% % 6th. author
% \alignauthor Charles Palmer\\
%        \affaddr{Palmer Research Laboratories}\\
%        \affaddr{8600 Datapoint Drive}\\
%        \affaddr{San Antonio, Texas 78229}\\
%        \email{cpalmer@prl.com}
}

\maketitle

%\onecolumn

%!TEX root = main.tex

\begin{abstract}
We consider the equity and fairness of curricula derived from Knowledge Tracing models. We begin by defining a unifying notion of an equitable tutoring system as a system that achieves maximum possible knowledge in minimal time for each student interacting with it. Realizing perfect equity requires tutoring systems that can provide individualized curricula per student. In particular, we investigate the design of equitable tutoring systems that derive their curricula from Knowledge Tracing models. We first show that many existing models, including classical Bayesian Knowledge Tracing (BKT) and Deep Knowledge Tracing (DKT), and their derived curricula can fall short of achieving equitable tutoring. To overcome this issue, we then propose a novel model, Bayesian-Bayesian Knowledge Tracing  (B\textsuperscript{2}KT), that naturally enables online individualization and, thereby, more equitable tutoring. We demonstrate that curricula derived from our model are more effective and equitable than those derived from classical BKT models. Furthermore, we highlight that improving models with a focus on the fairness of next-step predictions might be insufficient to develop equitable tutoring systems.
\end{abstract}

%% A category with the (minimum) three required fields
%%A category including the fourth, optional field follows...
%\category{D.2.8}{Software Engineering}{Metrics}[complexity measures, performance measures]

\keywords{equity \& fairness,
Bayesian Knowledge Tracing,
intelligent tutoring systems} % NOT required for Proceedings

%!TEX root = main.tex

\section{Introduction}

In recent years Massive Open Online Courses (MOOCs) and online educational platforms have gained significant importance.
%and online tutoring systems have gained importance in recent years.
They hold the opportunity of providing education at scale and making education accessible to a larger part of the world's population.
% educational resources broadly available by offering a low-cost possibility for scaling teaching activities to previously unseen levels.
To facilitate learning in online education and enable customized learning paths for all students, intelligent tutoring systems can be employed while limiting the amount of manual work necessary for each student~\cite{paviotti2012intelligent}.

In that context, moving education from an offline setting to an online setting has the potential to promote \textit{Inclusion, Diversity, Equity, and Accessibility (IDEA)}.
In particular, by reducing personnel efforts for tutoring, there is the opportunity to include students with diverse backgrounds and skills, and, importantly, to support their learning equitably.
%Moving to online education with automated support has the potential to promote IDEA
%
To achieve this, an intelligent tutoring system must be able to adapt to the specific characteristics of each student.

While individualized tutoring has been studied in the community for many years, we consider individualization with a focus on equitable and fair tutoring in this paper.
We start by providing a unifying definition of an equitable tutoring system.
Our definition is based on the ethical principles of \textit{beneficence} (``do the best'') and \textit{non-maleficence} (``do not harm'') which are commonly adopted in bioethics and medical applications~\cite{beauchamp2001principles} and strongly related to  recent work in the educational data mining field~\cite{doroudi2019fairer}.
Concretely, the principle of beneficence dictates that we should provide tutoring which maximizes the achieved knowledge.
The principle of non-maleficence dictates that this maximal knowledge should be achieved while minimizing a student's efforts.

In this paper, we particularly focus on how Bayesian Knowledge Tracing (BKT)~\cite{corbett1994knowledge} can be modified to better realize these ethical principles.
To this end, we propose the Bayesian-Bayesian Knowledge Tracing (\BBKT) model and demonstrate its advantages for equitable tutoring in several experiments.
Furthermore, we investigate the relation of the commonly considered AUC score concerning the derived tutoring policies, finding that even if a BKT model appears fair in terms of the AUC score, the derived tutoring policies can be inequitable.
This observation suggests that evaluating models concerning their AUC score can be insufficient to ensure equitable tutoring and that improving models with the sole focus on AUC might not result in more equitable tutoring.

In summary, we make the following contributions:
\vspace{-3mm}
\begin{itemize}
  \item We propose a unifying definition of equitable tutoring motivated by the ethical principles of beneficence and non-maleficence.

  \item We propose the (\BBKT) which allows for effective individualization and demonstrate its benefits concerning equitable tutoring. 
  
  \item We highlight that focusing on equity in terms of AUC can be insufficient to ensure equitable tutoring in terms of our definition.

\end{itemize}

\clearpage
%!TEX root = main.tex

\section{Related Work}
\label{sec:related}

\paragraph{Fairness in online education and BKT}

Several works have considered fairness in data-driven educational systems and intelligent tutoring, e.g., \cite{kizilcec2020algorithmic,doroudi2019fairer,yu2020towards,lee2020evaluation}.
In~\cite{kizilcec2020algorithmic}, the authors discussed implications of using data-driven predictive models for supporting education on fairness.
They identified sources of bias and discrimination in ``the process of developing and deploying these systems''.
Relevant to promoting IDEA, they discussed several high-level possibilities to improve fairness of systems in the ``action step'', e.g., through modifying predictions of models for groups of students with different characteristics.
It remains open how to operationalize such an approach in cases in which the required  knowledge of the group membership is not  available in the first place.
In~\cite{yu2020towards}, it was investigated whether different data sources can provide helpful information to predict students' success in education.
They found that different data sources (institutional data and learning management system data) can help to make better predictions but have different characteristics in whether they over- or underestimate students' success for historically disadvantaged subpopulations of students.
The authors of~\cite{lee2020evaluation} investigated fairness trade-offs when students' successes are predicted from university administrative records and identified gender and racial bias in some of the considered fairness measures. 
Through post-hoc adjustments they investigated how this problem can be alleviated and which trade-offs arise.
\cite{doroudi2019fairer} studied fairness in the context of BKT.
In particular, they showed that tutoring policies basing on BKT models can be inequitable when these models are inaccurate and that such inaccurate models can stem from fitting BKT models to populations of students or using models based on incorrect assumptions.
They consider the \textnormal{equity-gap} as a measure of unfairness, which is the difference between the percentage of fast and slow students mastering a skill.
%(more generally this can be seen as the difference in learning success for different subpopulations).

%In contrast we consider how quickly different models can reduce the equity gap by being adaptive to the students' parameters. 

\vspace{-6mm}
\paragraph{Individualization in BKT}

Several papers have studied individualization of BKT models per student, e.g., ~\cite{lee2012impact,pardos2010modeling,yudelson2013individualized}.
In~\cite{pardos2010modeling} the \emph{prior per student} model was introduced in which a student-specific parameter characterizing the students' individual knowledge is used.
The parameters are fitted using an EM procedure.
Importantly, the parameters for each student are fixed during inference by providing the student' identities while in our \BBKT model they are ``re-weighted'' according to the evidence from interaction.
Another example is \cite{yudelson2013individualized}, which considered individualization through defining student and skill specific parameters which are fitted through gradient descent. 
The models show tangible improvements in prediction performance of unseen students.

\vspace{-6mm}
\paragraph{Instructional policies}
A key necessity for achieving equity according to our definition is using instructional policies which stop practicing a skill at the right time.
This problem has for instance been considered in different contexts in~\cite{kaser2016stop,pelanek2018conceptual}.
In~\cite{kaser2016stop}, a stopping condition based on predictive stability, i.e., whether predictions about a student solving some task correctly are stable, was proposed.
The resulting policies are particularly useful for student models which allow forgetting and could be combined with our proposed \BBKT model.
%We leave an evaluation of this idea to future work.
Further related work has  investigated approaches for leveraging deep generative models for creating policies to quickly assess students' knowledge~\cite{wang2021QuestionMining} and using reinforcement learning for optimizing tutoring policies ~\cite{singla2021reinforcement,he2021quizzing}.

%!TEX root = main.tex

\section{Background \& Notation}

In this section, we introduce the necessary background and notation for the remainder of this paper.

\subsection{Bayesian Knowledge Tracing}

Bayesian knowledge tracing (BKT)~\cite{corbett1994knowledge} is a model characterizing the skill acquisition process of students.
For a single skill, it can be understood as a standard hidden Markov model (HMM) in which the binary latent state encodes the mastery of the skill, and the observations correspond to binary random variables indicating whether a practicing opportunity of the skill at a specific time was solved correctly or not.
%The mastery of a skill is also modeled as a binary random variable --- a skill is either mastered or not. 
% A student can practice skills through exercises and we can make binary observations indicating whether the student solved the exercise correctly or not.
Upon practicing the skill, the student acquires the skill with probability $p(T)$ if they have not already mastered the skill.
Once a skill is mastered, it remains mastered, i.e., there is no forgetting.
If a student has mastered the skill practiced by an exercise, they solve this exercise correctly with probability $1-p(S)$, where $p(S)$ is the probability of making a mistake.
If a student has not mastered the skill, it guesses the correct answer with probability $p(G)$.
%, i.e., $1-p(G)$ is the probability of not solving the exercise correctly.
% Typically it is assumed that once a skill is acquired, there is no forgetting, i.e., the skill is mastered for ever.
Before any interaction with the tutoring system, a student has already mastered the skill with probability $p(L_0)$.

In settings with multiple skills, these skills are often assumed to be acquired independently.
That is, if there are $\nskills$ skills, the skill acquisition process is modeled by $\nskills$ independent HMMs.

\subsection{Deep Knowledge Tracing}

In deep knowledge tracing~\cite{piech2015deep} the skill acquisition process is not modeled by an HMM as above but rather implicitly modeled through a recurrent neural network (RNN).
In particular, an RNN is trained to predict whether a student will answer the next exercise correctly based on input consisting of which exercises were practiced in the past and whether these practicing opportunities were solved successfully.
There is typically no explicit representation of the knowledge state of students in DKT. 
In the context of DKT previous work has mainly looked into so-called \textsc{Expectimax} policies which select exercises to practice in order to maximize the average probability that the next exercise will be solved correctly.
These policies can be computed for different lookaheads, e.g., for a one-step look-ahead in a greedy and myopic way or for longer-lookaheads using techniques for approximate planning or using exhaustive search.

\subsection{Notation}

We consider the interaction of students $s \in \students$ with an intelligent tutoring system.
The interaction history up to time $t$ is collected in a data set $\calD_t^s=\{ (z_1, c_1), (z_2, c_2), \ldots, (z_t, c_t)\}$, where $z_{t'} \in \skills$ is the skill practiced through an exercise at time $t'$, $c_{t'}\in \{0,1\}$ is an indicator of whether the exercise was solved correctly, and $\skills$ is the set of skills.
In the context of BKT, we refer to the random variables (RVs) indicating whether skill $i \in \skills$ is mastered at time $t$ as $Z_t^i$ and to the RVs indicating whether an exercise practicing that skill would be solved correctly as $C_t^i$.
Sometimes we use an additional superscript $s$ to indicate the student the random variables correspond to.
We use upper-case terms like $Z_t^i$ to denote RVs and their lower-case counterparts like $z_t^i$ to denote particular instantiations of these RVs.

%!TEX root = main.tex

\section{Operationalizing Equity}

In this section, we provide a definition of equity in intelligent tutoring and discuss its operationalization for promoting IDEA.
%in the context of BKT-models for EDM, and discuss relations to ethical principles commonly adopted in the field of machine learning. 

%%%%%%%%%%%%%%%%%%%%%%%%%%%%%%%%%%%%%%%%%%%%%%%%%%%%%%%%%%%%%%%%
%%%%%%%%%%%%%%%%%%%%%%%%%%%%%%%%%%%%%%%%%%%%%%%%%%%%%%%%%%%%%%%%
%%%%%%%%%%%%%%%%%%%%%%%%%%%%%%%%%%%%%%%%%%%%%%%%%%%%%%%%%%%%%%%%
\subsection{Setting}

We consider a tutoring setting in which a total of $K$ sills ought to be taught.
These skills are taught by an intelligent tutoring system that employs a tutoring policy $\pi\colon \history\rightarrow \items \cup \{ \terminate \}$.
This policy maps histories $h \in \history$ consisting of (previous) observations of a student's learning process to an exercise $e \in \items$ to be practiced next or to a stop-action $\terminate$, which ends the teaching process.
%, i.e., the tutoring system does not suggest to conduct any further exercises.
We collect all tutoring policies $\pi$ of the above form into the set $\policies$.

We assume interaction with a set of students $\students$.
The students can have different characteristics in terms of their learning behavior.
That is, if the learning behavior of the students can be characterized by BKT models, the corresponding parameters $p(L_0)$, $p(S)$, $p(G)$ and $P(T)$ can differ for different students.
Every tutoring policy $\pi$ results in an expected stopping time $T^s(\pi)$, i.e., the expected time of executing the stop action, and an expected knowledge $L^s(\pi)$ acquired by the end of the teaching process, i.e., $L^s(\pi)$ is the expected number of mastered skills when the stop action is executed.
% A teaching policy $\pi$ pareto-dominates another teaching policy $\pi'$ if $T^s(\pi) \leq T^s(\pi')$ and $L^s(\pi) \geq L^s(\pi')$.

For defining a notion of equity in this setting, we turn to the ethical principles of \textit{beneficence} and \textit{non-maleficence}.
We understand them to translate into the objective of maximizing a student's knowledge (``do the best'') using as little of the student's resources as possible (``do not harm''), i.e., performing a minimal number of exercises:
\begin{definition}
%[Equitable tutoring system]
  \label{def:equity}
  Consider a tutoring system employing a tutoring policy $\pi \in \Pi$.
  The tutoring policy $\pi$ is equitable for student $s$ if and only if
  \begin{align*}
      T^s(\pi) = \min_{\pi' \in \policies, L^s(\pi')=K} \; T^s(\pi') \quad \textnormal{ and } \quad L^s(\pi) = K,
  \end{align*}
  where $K$ is the total number of skills.
  A tutoring system is equitable if its tutoring policy is equitable for all students $s \in \students$.
\end{definition}
Thus, informally, a tutoring system is equitable if it can teach all $K$ skills in the minimal possible amount of time to any student.
Please note that our notion of equity is strongly related to that introduced in~\cite{doroudi2019fairer} (cf.\ discussion below).
% and Section~\ref{sec:related}

In the above definition, we assume that all students can master all $K$ skills.\footnote{Otherwise, if a student existed that was not able to acquire all skills, no tutoring system would be equitable according to Definition~\ref{def:equity}.
Our definition can be generalized in a straightforward manner to account for an individual student's maximal achievable knowledge.}
Importantly, a tutoring system can only be equitable if it is adaptive to the students which are interacting with it. 
In particular, it has to individualize the assignment of exercises and needs to carefully select the "stop action", in order to achieve equity.
The above definition describes an idealized notion of equity which in general cannot be achieved as the tutoring policy would have to be informed by a kind of ``oracle'' in order to teach using the optimal policy within $\Pi$ right from the beginning.

Nevertheless, we can compare tutoring policies $\pi$ in the spirit of the above definition. 
In particular, given two tutoring policies $\pi$ and $\pi'$ which both teach the same number of skills, we consider the policy $\pi$ to be \emph{more equitable} as compared to $\pi'$ if for all students $s \in \students$ it holds that
%the following holds:
$T^s(\pi) \leq T^s(\pi').$

We note that our notion of equity is strongly related to that introduced in~\cite{doroudi2019fairer}.
In~\cite{doroudi2019fairer}, the authors ``assume that an equitable outcome is when students from different demographics reach the same level of knowledge after receiving instruction''.
As such, there is no need to reach this level of knowledge fast; hence this basic notion of equity is in principle trivial to achieve by providing vast amounts of exercises (as also noted by the authors of~\cite{doroudi2019fairer}). %, such an approach is useless in practice.
The desideratum of achieving knowledge fast is later added on top of their notion of equity whereas in our case it is a fundamental constituent in the first place.
Furthermore, our interest extends to downstream implications of such a definition of equity, namely the individualization of knowledge tracing.

%%%%%%%%%%%%%%%%%%%%%%%%%%%%%%%%%%%%%%%%%%%%%%%%%%%%%%%%%%%%%%%%
%%%%%%%%%%%%%%%%%%%%%%%%%%%%%%%%%%%%%%%%%%%%%%%%%%%%%%%%%%%%%%%%
%%%%%%%%%%%%%%%%%%%%%%%%%%%%%%%%%%%%%%%%%%%%%%%%%%%%%%%%%%%%%%%%
\subsection{Theoretical Implications}

Our definition of equity immediately leads to the (probably obvious) but important observation, that equity of a tutoring system can in general only be achieved by individualization:
%of the tutoring policy:
\begin{observation}
  A tutoring system for a population of students with different learning characteristics can only be equitable if its tutoring policy is adaptive to the students.
  % Non-optimal model $\Leftrightarrow$ non-equitable outcome
\end{observation}
\vspace{-2mm}
Thus, we note that if the tutoring policy is derived deterministically from a non-adpative model of the students, the tutoring system will in general not be equitable.

Furthermore, as already mentioned, our definition of equity is in general impossible to achieve.
In particular, achieving equity would require basing a policy on rich side information in order to employ an optimal tutoring policy for any particular student right from the beginning.
Such side information might not be available or insufficient to determine the optimal policy.

%%%%%%%%%%%%%%%%%%%%%%%%%%%%%%%%%%%%%%%%%%%%%%%%%%%%%%%%%%%%%%%%
%%%%%%%%%%%%%%%%%%%%%%%%%%%%%%%%%%%%%%%%%%%%%%%%%%%%%%%%%%%%%%%%
%%%%%%%%%%%%%%%%%%%%%%%%%%%%%%%%%%%%%%%%%%%%%%%%%%%%%%%%%%%%%%%%
\subsection{Operationalization}

In practice, tutoring policies are often either simple fixed strategies, e.g., repeating exercising a skill until a certain number of consecutive exercises related to that skill were solved successfully, or derived from a model, e.g., a BKT model, such that each knowledge component is repeatedly exercised until it is mastered with a certain probability.

Tutoring policies based on incorrect or non-adaptive models can result in a student not acquiring all skills or suggest to perform too many practicing opportunities.
Thus the following two general directions are important for building equitable tutoring systems:
\vspace{-3mm}
\begin{enumerate}
    \item \textbf{Using side information.}
      To start the interaction with a student as effectively as possible, any available side information about a student should be used to individualize the underlying models.
      For instance, in the context of RNN based DKT models, the side information could be used to predict the initial hidden states of the RNNs before the interaction with the student starts.
      In the context of classical BKT models, the side information could be used to make an initial guess about the key parameters of the model (initial knowledge, transition, slip and guess probabilities).
      Unfortunately, we are not aware of publicly available datasets which come with such side information and therefore it is difficult to develop such approaches.
      %, cf.~Section Outlook.
    
    \item \textbf{Online adaptation.}
      Even when using side information, a model is likely not perfectly individualized to all students it encounters.
      To further adjust the models in such cases, online adaption of the models during interaction seems promising.
      This can for instance be operationalized by training models for different sub-populations of students and during usage decide on which model is best fitted for a particular student.
      An alternative approach, which we propose in this paper, is a Bayesian BKT model, which performs posterior parameter inference about students' properties and leverages the results for further predictions.
      %, cf.~Section~\ref{sec:models}.
\end{enumerate}

%\input{existing.tex}
%!TEX root = main.tex

\section{Bayesian-Bayesian Knowledge\\ Tracing}
\label{sec:models}

\begin{figure}
    \centering
    \includegraphics[width=.75\columnwidth]{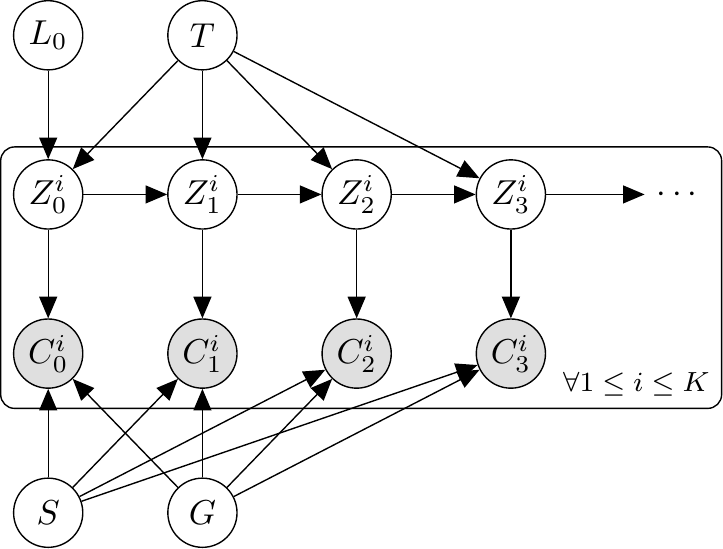}
    \caption{Representation of Bayesian-Bayesian Knowledge Tracing as a probabilistic graphical model.
    The acquisition and application of $\nskills$ skills is controlled by global parameters $p(L_0)$, $p(S)$, $p(G)$, and $p(T)$, i.e., they apply to all $\nskills$ skills. 
    %Also the skip and guess parameters $S$ and $G$ are global.
    }
    \label{fig:bbkt}
\end{figure}

As outlined in the previous section, an equitable tutoring system must employ policies that are adaptive to students' properties.
As these properties might not be known before actually interacting with a student, online adaption during interaction is important.
To this end, we propose a Bayesian variant of the classical BKT model from which individualized policies can be derived, cf.\ Figure~\ref{fig:bbkt}.

In particular, we assume that each student $s$ has its own learning dynamics, described by student-specific parameters $\theta^s$.
If the learning dynamics can be described using a BKT model, which we assume in the following, $\theta^s=(p(L_0^s), p(T^s), p(S^s),p(G^s))$.
We assume these learning dynamics to apply for the acquisition of \emph{all} skills, i.e., they are \emph{global} parameters of a student.
If we knew these parameters for a student, we could use them in a classical BKT model and would likely observe good tutoring performance.
% when using a suitable tutoring policy
However, in practice, we don't know these parameters and need to infer them. 
To this end, we take a Bayesian approach.
Therefore, we assume a set of possible parameters $\Theta$ such that $\theta^s \in \Theta$ and a prior distribution $p_0(\theta^s)$.
Based on $t$ observations of a student practicing exercises collected in data set $\calD_t$, we can compute the probability that a student has mastered a specific skill and base tutoring policies thereon.
Note that as we don't know $\theta^s$, this requires marginalizing out the (unknown) parameters $\theta^s$.
In this way the different possible parameters and their influence for predicting the knowledge state get re-weighted according to the available data.
In particular, we need to compute
\begin{align}
  p(Z_t^{s,i} \mid \calD_t) = \int_{\theta \in \Theta} \underbrace{p(Z_t^{s,i} \mid \theta, \calD_t)}_{=: (\#1)} \underbrace{p(\theta \mid \calD_t)}_{=: (\#2)} \textnormal{d}\theta,
  \label{eq:posterior}
\end{align}
where $Z_t^{s,i}$ is a random variable indicating whether skill $i$ is mastered at time $t$ by student $s$.
For only a few possible parameters $\theta$, the above equation can be solved exactly by enumeration and by observing that both terms $(\#1)$ and $(\#2)$ can be computed efficiently by leveraging the forward algorithm through the following recursion:
\begin{align*}
    \alpha_0^\theta(1) &= p(Z_0^{s,i} = 1 \mid \theta) = p(L_0) \\
    \alpha_0^\theta(0) &= p(Z_0^{s,i} = 0 \mid \theta) = 1 - p(L_0) \\
    \alpha_{t+1}^\theta(l) &= p(Z_{t+1}^{s,i}=l, \c^i_{t+1}) = \sum_{z_t^{s,i}} p(Z_{t+1}^{s,i}=l,Z_{t}^{s,i}=z_t^{s,i}, \c^i_{t+1}) \\
      &= \sum_{z_t^{s,i}} p(c^i_{t+1} \mid Z_{t+1}^{s,i}=l) p(Z_{t+1}^{s,i} = l \mid Z_{t}^{s,i} = z_{t}^{s,i}) \alpha_{t}(z_t^{s,i})
\end{align*}
where $\c^i_t$ collects all observations with respect to practicing the $i$th skill up to time $t$, and $c_{t'}^i$ is the $t'$th entry of $\c^i_t$.
Then
\begin{align*}
    (\#1) &= p(Z_t^{s,i} = 1 \mid \theta, \calD_t) = \frac{\alpha_t^\theta(1)}{\alpha_t^\theta(0) + \alpha_t^\theta(1)}, \textnormal{ and} \\
    (\#2) &= p(\Theta = \theta \mid \calD_t) = \frac{p_0(\theta) \cdot (\alpha_t^\theta(0) + \alpha_t^\theta(1))}{\sum_{\theta' \in \Theta} p_0(\theta') \cdot (\alpha_t^{\theta'}(0) + \alpha_t^{\theta'}(1))}.
\end{align*}
If $\Theta$ is large/continuous MCMC-techniques can be used to approximate Equation~\eqref{eq:posterior}.

In our experiments, we mainly consider a concrete instantiation of our proposed model in which we consider the learning rate to be unknown. 
This is motivated by previous work which has identified the learning rate as a key parameter for improving BKT based models in terms of AUC~\cite{yudelson2013individualized}.

\vspace{-3mm}
\paragraph{Advantages of \BBKT}
Through posterior parameter inference based on observed evidence, the proposed model can often yield (approximately) equitable tutoring policies.
%which are more equitable than those of classical BKT models with incorrect parameters.
In contrast, classical BKT models cannot adjust to the students' properties, possibly resulting in inequitable policies.

Furthermore, Bayesian treatment of the parameters can enable us to leverage ideas from active learning to actively seek information about students with the goal of teaching more effectively later.
This could be a clear advantage over other approaches for individualization, e.g., those using EM algorithms --- in such a case it is not obvious how to actively seek information about a students' parameters.
We did not explore this idea in this paper but consider this as a promising direction for future work.

%!TEX root = main.tex

\section{Experiments}

In this section, we compare tutoring policies derived from \BBKT, BKT, and DKT models in terms of equity.

\subsection{Experimental Setups}

In all presented results we denote the average stopping time of a policy for a population of students by $T_{\textnormal{stop}}$ and the average number of acquired skills by $\% \textnormal{ skills}$.

\vspace{-3mm}
\paragraph{Curricula}

%We consider curricula derived from the proposed B\textsuperscript{2}KT model and several baselines:
We consider the following curricula:
%and the inherent trade-offs by parametric choices, e.g., the mastery threshold in BKT.
%\paragraph{Curricula}
%In our experiments we consider the following policies:
\vspace{-3mm}
\begin{itemize}

    \item \textsc{Threshold}($\tau$):
      These curricula are based on knowledge tracing algorithms which allow computing the posterior probability of skill mastery.
      In particular, these curricula repeatedly exercise a skill until it is mastered with a probability of at least $\tau$.

    \item \expectimaxPolicy{1}:
      These curricula are derived from DKT models which do not enable us to compute a posterior probability of skill mastery.
      Skills that maximize the average probability that a practicing opportunity for a randomly selected skill at the next step would be solved correctly are practiced.
      The stop action is invoked when the predicted probabilities change by less than a threshold from one step to the next.
      
\end{itemize}

\vspace{-7mm}
\paragraph{Models}

We consider the following models:
\vspace{-3mm}
\begin{itemize}
    \item \textsc{BKT}: the classical BKT models with fixed parameters.
    \item \textsc{B\textsuperscript{2}KT}: the proposed Bayesian BKT model.%, cf.~Section~\ref{sec:models}.
    \item \textsc{DKT}: deep knowledge tracing models~\cite{piech2015deep} based on the implementation provided by~\cite{gervet2020deep}. 
\end{itemize}
\vspace{-2mm}

\begin{table*}[!ht]
    \caption{Equity trade-offs of curricula derived from different models/parameterizations.}
    \label{tab:curricula}
    % (\texttt{experiment-tradeoff.py})
    \centering
    \scalebox{0.90}{
    \begin{tabular}{ccccccccccccc}
      \toprule
%         & \multicolumn{6}{c}{\# skills} \\\cmidrule{2-7}
                                       & \multicolumn{4}{c}{1 skill} & \multicolumn{4}{c}{5 skills} & \multicolumn{4}{c}{20 skills} \\\cmidrule(lr{.75em}){2-5} \cmidrule(lr{.75em}){6-9} \cmidrule(lr{.75em}){10-13} 
                                       & \multicolumn{2}{c}{slow learners} & \multicolumn{2}{c}{fast learners}    & \multicolumn{2}{c}{slow learners} & \multicolumn{2}{c}{fast learners}  & \multicolumn{2}{c}{slow learners} & \multicolumn{2}{c}{fast learners} \\\cmidrule(lr{.75em}){2-3} \cmidrule(lr{.75em}){4-5} \cmidrule(lr{.75em}){6-7} \cmidrule(lr{.75em}){8-9} \cmidrule(lr{.75em}){10-11} \cmidrule(lr{.75em}){12-13} 
        \thresholdPolicy{0.95}  & \% skills & $T_{\textnormal{stop}}$  & \% skills & $T_{\textnormal{stop}}$ & \% skills & $T_{\textnormal{stop}}$ & \% skills & $T_{\textnormal{stop}}$ & \% skills & $T_{\textnormal{stop}}$ & \% skills & $T_{\textnormal{stop}}$  \\\midrule
        BKT slow      & 97.00 & 24.14 & 99.50 & 9.49        & 97.20 & 122.80 & 99.90 & 66.00    & 97.55 & 492.64 & 99.90 & 183.84 \\[0.3em]
        BKT fast      & 61.00 & 13.85 & 97.50 & 5.96        & 62.60 & 71.98  & 96.10 & 29.81    & 64.20 & 288.76 & 97.23 & 120.59 \\[0.3em]
        BKT mixed     & 95.00 & 23.51 & 100.00 & 8.33       & 95.40 & 113.67 & 99.90 & 40.93    & 94.53 & 466.55 & 99.68 & 169.86 \\[0.3em]
        \BBKT         & 94.50 & 24.04 & 100.00 & 7.88       & 97.70 & 120.87 & 98.40 & 32.61    & 96.68 & 493.00 & 96.66 & 120.05\\\bottomrule
    \end{tabular}
    }
    \vspace{-3mm}

    % \sebastian{Show a plot where the mastery threshold is varied.}
\end{table*}

\subsection{Experimental Results}

%%%%%%%%%%%%%%%%%%%%%%%%
%%%%%%%%%%%%%%%%%%%%%%%%
%%%%%%%%%%%%%%%%%%%%%%%%

\paragraph{Students with different learning behaviors}
\vspace{-4mm}
We study the equity of tutoring policies when the students are sampled uniformly from two groups, each containing students with learning dynamics described by a ground truth BKT model.
In particular, we build on the experimental setup from~\cite{doroudi2019fairer} where there is a group of slow learners (BKT slow) and fast learners (BKT fast).
In~\cite{doroudi2019fairer}, the authors also fitted a BKT model to interaction data from students from both groups; we refer to the corresponding BKT model as BKT mixed.
The parameters of these models are as follows:
\vspace{-4mm}
\begin{center}
\begin{tabular}{c|cccc}
             & $p(L_0^s)$ & $p(S^s)$ & $p(G^s)$ & $p(T^s)$ \\\hline
     BKT slow & $0.0$    & $0.2$  & $0.2$  & $0.05$ \\
     BKT fast & $0.0$    & $0.2$  & $0.2$  & $0.3$ \\
     BKT mixed & $0.071$ & $0.203$ & $0.209$ & $0.096$
\end{tabular}
\end{center}
\vspace{-4mm}

\begin{figure}
    \centering
    \includegraphics[scale=.9]{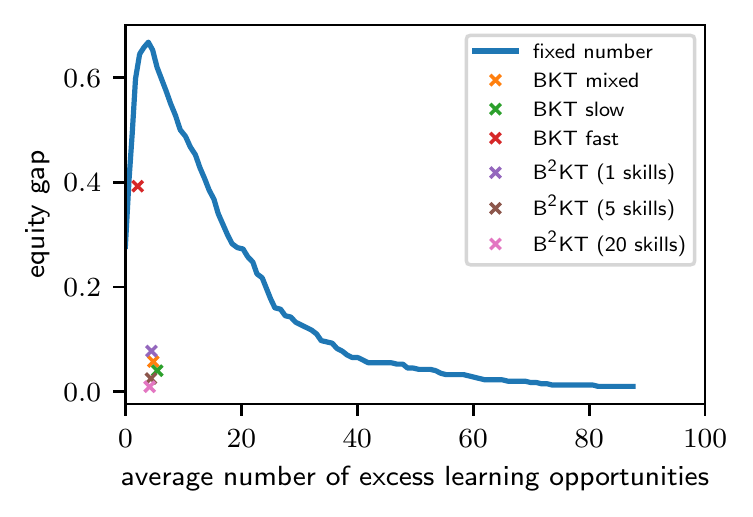}
    \vspace{-3mm}
    \caption{Equity gap versus number of excess learning opportunities for different policies. We use a \thresholdPolicy{0.95} tutoring policy derived from the knowledge tracing models.
    \BBKT becomes more equitable as more skills are taught.}
    % \texttt{experiment-fairness.py}
    \label{fig:fairness-over-skills}
    \vspace{-3mm}
\end{figure}

We considered the interaction with 400 students, 200 from each group,
and we compared the performance of \textsc{Threshold}(0.95) tutoring policies based on the different BKT models for different numbers of skills that ought to be taught in Table~\ref{tab:curricula}.
We observe that in the case of mismatch of the student properties and the BKT models used for the threshold policy, either only a small fraction of the skills (clearly below 95 \%) is acquired or that more than necessary time is spent exercising. 
Furthermore, in order to teach the desired 95 \% of the skills to slow students when basing the policy on the BKT model for fast students, the threshold $\threshold$ would have to be increased significantly, thereby providing unnecessary extra exercises to fast students.
For example, by setting the threshold $\threshold$ to $0.998$, the fraction of skills learned by the slow learners increased to $0.95$ and the stopping time to $T_\textnormal{stop}=464.08$ (much worse as compared to the \thresholdPolicy{0.95} policy for the correct model).
At the same time, the stopping time for the fast learners increased to $T_\textnormal{stop}=174.72$.
%%% experiment-tradeoff2.py, experiment3
The mismatch issue is alleviated in the case of the \BBKT~model (assuming a uniform prior over both types of students), in particular for a larger number of skills.
Intuitively this is because, in the case of multiple skills, the model has more opportunities to learn about the students' properties and use these insights in later tutoring.
This is an important characteristic of a model with respect to equity as the model becomes ``more equitable'' as it learns about the students. % to the whole student population
This fact is also illustrated in Figure~\ref{fig:fairness-over-skills} in which we reproduce and extend an experiment from~\cite{doroudi2019fairer} in which they compare the ``equity gap'' (the difference in the percentage of skills mastered by fast students to the percentage of skills mastered by slow students) to the number of excess learning opportunities.
In particular, we extended the figure to the case of teaching multiple skills and show the performance of \BBKT. 
Importantly, \BBKT~becomes more equitable (the corresponding point moves to lower left) as more skills are taught.

%%%%%%%%%%%%%%%%%%%%%%%%
%%%%%%%%%%%%%%%%%%%%%%%%
%%%%%%%%%%%%%%%%%%%%%%%%

\vspace{-5mm}
\paragraph{Out-of-distribution generalization}

We evaluate whether \BBKT can help with aspects relevant to \emph{inclusion and diversity}, the first two principles of IDEA. 
In particular, we consider a stylized mismatch setting in which a tutoring system interacts with students who have a learning behavior that was not encountered in the data that was used for constructing the BKT models.
%informing the threshold policies.
In particular, in addition to the previous two types of students, we assume a third type of learner (BKT med) with %learners faster than slow learners but slower than fast learners.
the following parameters:  $p(L_0^s)=0.0,p(S^s)=0.2,p(G^s)=0.2,p(T^s)=0.18$.
% $s$ are parameters for simulating such a students
We considered \thresholdPolicy{0.95} policies based on BKT models of slow and fast learners and the \BBKT model with a uniform prior over slow and fast learners.
Our results are presented in Table~\ref{tab:ood}.
We observe that the performance of the policies derived from the \BBKT model have comparable performance to those derived from the true model
%with an increasing number of skills
whereas other models yield policies clearly worse in terms of stopping at the right time or teaching the right amount of skills.
Although the correct model has 0 prior probability (and thus also zero posterior probability) in the \BBKT model, the model performs well.
Such a characteristic can be of key importance for promoting inclusion, e.g., when interacting with students who were not represented in a data set that was used for building an intelligent tutoring system or students who were underrepresented in that data.

%%%%%%%%%%%%%%%%%%%%%%%%
%%%%%%%%%%%%%%%%%%%%%%%%
%%%%%%%%%%%%%%%%%%%%%%%%
\vspace{-5mm}
\paragraph{Fair next step predictions do not necessarily imply equitable tutoring}
We show empirically that models which might appear  fair when looking at their AUCs for different groups of students do not necessarily yield equitable tutoring policies.
In particular, we again focus on a student population consisting of two groups of learners:%\\[3mm]
\vspace{-4mm}
\begin{center}
\begin{tabular}{c|cccc}
             & $p(L_0^s)$ & $p(S^s)$ & $p(G^s)$ & $p(T^s)$ \\\hline
     Group 1 & $0.0$    & $0.1$  & $0.4$  & $0.1$ \\
     Group 2 & $0.0$    & $0.1$  & $0.2$  & $0.3$ \\
\end{tabular}
\end{center}
\vspace{-4mm}
%\\[3mm]
We generated data for 400 students in a setting with 20 skills and 1000 random exercises from BKT models, where $50\%$ of the students are from group 1 and group 2, respectively.
The true model of group 1's students achieved an AUC of $0.7393$ for group 1's students, while the true model of group 2's students achieved an AUC of $0.6710$ for group 2's students.

Looking only at the AUC, the two models appear rather inequitable (there is no group parity).
Thus it might appear sensible to aim to use a BKT model for tutoring which has comparable AUCs for both groups in order to promote equity.
For instance, a BKT model using parameters  $p(L_0)=0$, $p(S)=0.4$, $p(G)=0.1$, $p(T)=0.65$ achieves an AUC of $0.6719$ on group 1's students and of $0.6733$ on group 2's students, respectively.
That is, the AUCs on the two groups are approximately equal.
%and in this sense, the model appears fair.

However, when looking at the different models with respect to the tutoring performance when using a \textsc{Threshold}($0.95$)-policy, we observe a very different picture.
We summarize our findings in Table~\ref{tab:AUCfairness}.
In particular, the fraction of skills taught differs significantly between the two groups:
In group 1 only $28.68 \%$ of the skills are acquired by the students on average while in group 2 $74.70 \%$  of the skills are acquired on average.
This finding is closely related to the observation that models with greatly different characteristics can have similar AUCs~\cite{pelanek2018details}.
%, and thus, that considering the AUC only might lead to incorrect conclusions about models \cite{pelanek2018details}.

In this context, we also evaluate \expectimaxPolicy{1} policies based on DKT models (trained using the implementation provided by~\cite{gervet2020deep}; using default settings for hidden sizes, etc.). 
The AUCs of the DKT model are 0.6757 and 0.6551 for slow and fast students respectively and so roughly comparable to that of the BKT models appearing to be equitable in terms of their AUC.
However, the performance of \expectimaxPolicy{1} policies we observed was very low (although the average predicted probability for answering questions correctly increased monotonically in most cases).
In particular, for a threshold $\tau=0.001$ it achieved a surprisingly low performance of 8.3 $\% \textnormal{ skills}$ / $T_\textnormal{stop}=41.01$ (slow learners) and 10 $\% \textnormal{ skills}$ / $T_\textnormal{stop}=29.80$ (fast learners). 
Interestingly, a difference in stopping time for slow and fast learners could be observed despite the model making overall bad recommendations.
When looking into this issue in more detail we observed that the policies repeatedly taught the same skill, i.e., DKT  did not correctly model the learning process---it appeared to only have picked up that the probability of answering exercises correctly increases over time.

\begin{table}[!htb]
    \caption{Out-of-distribution generalization.}
    \label{tab:ood}
    % (\texttt{experiment-tradeoffspy})
    \centering
    \scalebox{0.75}{
    \begin{tabular}{ccccccc}
      \toprule
%         & \multicolumn{6}{c}{\# skills} \\\cmidrule{2-7}
                                       & \multicolumn{2}{c}{1 skill} & \multicolumn{2}{c}{5 skills} & \multicolumn{2}{c}{20 skills} \\\cmidrule(lr{.75em}){2-3} \cmidrule(lr{.75em}){4-5} \cmidrule(lr{.75em}){6-7} 
                                       & \multicolumn{2}{c}{BKT med} & \multicolumn{2}{c}{BKT med}    & \multicolumn{2}{c}{BKT med} \\\cmidrule(lr{.75em}){2-3} \cmidrule(lr{.75em}){4-5} \cmidrule(lr{.75em}){6-7}
        \thresholdPolicy{0.95}  & \% skills & $T_{\textnormal{stop}}$  & \% skills & $T_{\textnormal{stop}}$ & \% skills & $T_{\textnormal{stop}}$ \\\midrule
        BKT slow      & 99.50 & 11.28 & 99.55 & 56.45 & 99.59 & 225.25\\[0.3em]
        BKT fast      & 90.75 & 7.61  & 91.55 & 37.59 & 91.70 & 151.36 \\[0.3em]
        BKT mixed     & 99.50 & 10.46 & 99.00 & 51.71 & 99.21 & 211.29 \\[0.3em]
        BKT med       & 98.25 & 8.82  & 97.35 & 45.59 & 97.84 & 184.20 \\[0.3em]
        \BBKT         & 98.75 & 10.33 & 97.50 & 48.80 & 94.19 & 168.36\\\bottomrule
    \end{tabular}
    }

%%%%%%%%%%%%%%%%%%%%%%%%%%%%%%%%%%%%%%%%

    \caption{Fairness in terms of similar AUCs on different groups does not imply fairness in terms of the derived teaching curricula. See the main text for details.}
    \label{tab:AUCfairness}
    
    \scalebox{0.9}{
    \begin{tabular}{ccccccc}
       \toprule 
               & \multicolumn{3}{c}{group fair wrt AUC} & \multicolumn{3}{c}{true model wrt group} \\\cmidrule(lr{.75em}){2-4} \cmidrule(lr{.75em}){5-7}
       group   & AUC & \% skills & $T_{\textnormal{stop}}$ & AUC & \% skills & $T_{\textnormal{stop}}$ \\ \midrule         
       group 1 & 0.6719 & 28.68 & 61 & 0.7393 & 96.13 & 308  \\
       group 2 & 0.6733 & 74.70 & 64 & 0.6710 & 96.35 & 105 \\\bottomrule
    \end{tabular}
    }

    % \note{\texttt{experiment-group-fairness.py}}
\end{table}

% \input{outlook.tex}
%!TEX root = main.tex

\section{Conclusions \& Future Work}

We considered the equity and fairness of curricula derived from knowledge tracing models. 
We provided a unifying definition of an equitable tutoring system motivated by the ethical principles of beneficence and non-maleficence.
Our definition is, in many practical settings, not realizable but suggests that the individualization of tutoring policies to students is key for realizing equity.
We proposed the \BBKT model, a Bayesian variant of the classical BKT model, and demonstrated in various experiments that it can be beneficial for realizing equitable tutoring systems and promoting IDEA more generally.
Furthermore, we highlighted that developing models with the sole focus on (fair) next-step predictions might be insufficient to develop equitable tutoring systems.
In future work we aim to build novel models and policies which allow individualization through leveraging side information and improve scalability of \BBKT by approximate inference techniques.

%ACKNOWLEDGMENTS are optional
%\section{Acknowledgments}
%This section is optional; it is a location for you
%to acknowledge grants, funding, editing assistance and
%what have you.  In the present case, for example, the
%authors would like to thank Gerald Murray of ACM for
%his help in codifying this \textit{Author's Guide}
%and the \textbf{.cls} and \textbf{.tex} files that it describes.

\clearpage

\bibliographystyle{abbrv}
\bibliography{sigproc}  % sigproc.bib is the name of the Bibliography in this case

\balancecolumns
% That's all folks!

\end{document}